\documentclass[journal]{IEEEtran}
\usepackage{blindtext}
\usepackage{graphicx}
\usepackage{subcaption}
\usepackage[utf8]{inputenc}
\usepackage[export]{adjustbox}
\usepackage{wrapfig}

\usepackage{amsmath}
\usepackage{amsfonts}
\usepackage{amssymb}

\begin{document}

\title{Autonomous Ingress of a UAV through a window using Monocular Vision}

\author{Abhinav Pachauri$^{1}$ ,Vikrant More$^{2}$,Pradeep Gaidhani$^{2}$,Nitin Gupta$^{2}$\\ $^{1}$ IIT Kharagpur , $^{2}$ Navstik Autonomous Systems, Pune 
}

\maketitle

\begin{abstract}

The use of autonomous UAVs for surveillance purposes and other reconnaissance tasks is increasingly becoming  popular and convenient.These tasks requires the ability to successfully ingress through the rectangular openings or windows of the target structure.In this paper , a method to robustly detect the window in the surrounding using basic image processing techniques and efficient distance measure, is proposed.Furthermore, a navigation scheme which incorporates this detection method for performing navigation task has also been proposed.The whole navigation task is performed and tested in the simulation environment GAZEBO .

\end{abstract}


\IEEEpeerreviewmaketitle

\section{Introduction}

The field of autonomous UAVs is still in its embryonic stage.The concept of advanced autonomous robotics is quite often depicted in science fiction movies and many of these are becoming a reality.These autonomous UAVs can serve the purpose such as search and rescue , evacuations , intelligence gathering and other reconnaissance task without much human intervention.This paper focuses on those tasks which require visual detection for navigation such as Ingress through rectangular window to enter into the building , landing and mapping tasks etc.A visual approach for detecting rectangular opening or window as a target in the surrounding has been proposed.  \\

Highly power-intensive sensor suites are difficult to use on UAVs for Ingress task.These sensors should be light, small and capable enough to perform the required tasks.In this work, single front facing monocular camera has been used for detection. An optical flow sensor is also mounted in order to get local positions accurately.

\section{Literature Review}

Different approaches have been used for similar work. Feature and pose constrained visual aided inertial navigation for computationally constrained aerial vehicles after detection of Ingress target (Window or door)has been presented in this paper[1], feature extraction and matching has been used for detection purposes.Feature detection is implemented by STAR feature detectors.Correspondence between detected frames are found using upright SURF( Speeded Up Robust Features).

Stereo vision system for the visual feed and  Simultaneous Localization and Mapping (SLAM) algorithm are also being used for the navigation purpose[2].Both the EKF state estimates and incoming sensor measurements are used to create a global map of the surrounding.
Both the above approaches use two cameras and highly power intensive sensors. 

The work of R. Brockers $et al.$[3] uses a single monocular camera to detect the navigation target in a known environment mapped using VICON .3D waypoints are being generated in the same frame for autonomous navigation.

Pose-based filtering [5], referred to as Vision Aided Inertial Navigation (VAIN), is in many respects similar to bundle adjustment. VAIN systems increases the accuracy of the filter state with current estimate of the camera position.It builds up a set of previous camera poses over time in order to get more accurate results. The complexity of filter is linear in the observed number of features which are stored among the past camera poses which are used to update the vehicle state.

Visual aided autonomous Navigation in the GPS-denied environment has also been explored . A vision-aided inertial navigation system that enables autonomous flight of an aerial vehicle in GPS-denied environments has been discussed by Allen D $et al.$ [6].Feature point information from a monocular vision sensor is used to bound the drift resulting from integrating accelerations and angular rate measurements from an Inertial Measurement Unit (IMU) forward in time. An Extended Kalman filter framework is proposed for performing the tasks of vision-based mapping and navigation separately.

Blosch et al.[7] discuss the use of a monocular camera-based SLAM algorithm for accurate estimation of the pose of a UAV in GPS denied environments.A technique proposed by Klein and Murray[4] for augmented reality applications is used by Blosch et al. to show the use of a SLAM algorithm for UAV real-time application with sustained tracking using image keyframes .Poses are generated externally for navigation task.This paper uses estimation of the relative poses for correcting the relative position of UAV.

\renewcommand{\thesubsection}{\thesection.\alph{subsection}}
\renewcommand{\thesubsubsection}{\thesubsection.\alph{subsubsection}}
\section{ Approach }
Target can be in any plane of the NED frame and the absolute yaw of the quadrotor is known from the Attitude estimator in the NED frame.So ,for getting the relative angle between the quadrotor(camera plane) and the target plane , homographic estimation is used. These relative poses will be used in the navigation scheme.
The task can be divided into different intermediate steps .These steps are described in details in this section.
\subsection{\textbf{Robust Window Detection}}
Robust window detection is a major step which should be done perfectly in order to get the camera pose in the next step correctly.Window detection step is further divided into the sub-steps which should all described below.
\subsubsection{\textbf{Edge detection}}
Standard Canny Edge Detection algorithm is applied for getting the edges.Some intermediate steps being inserted for more robust edge detection which includes removing small and broken edges, enhancing the target edges and normalization/equalization of Histogram(normalization of distribution of intensities among pixels). 
Steps for the Noise removal:
\begin{itemize}
\item Gaussian Blurring
\item Pyramid Downscaling/Up-scaling
\item Equalizing Histogram
\end{itemize}
First two steps are majorly for the blurring/smoothing the image as much as possible unless the window edges are affected because the edge detection is done using high pass filters and the signal part in the image should be lowered in order to get the edges correctly. Without any noise removal the edge detection can be seen in figure.1(a).The effect of the noisy edge detection on the final window detection can be seen in the figure.2(a) which is not satisfactory.
\\
\\
\begin{figure}[h]
\begin{subfigure}{0.2\textwidth}
\includegraphics[width=1.7in,height=1.25in,keepaspectratio]{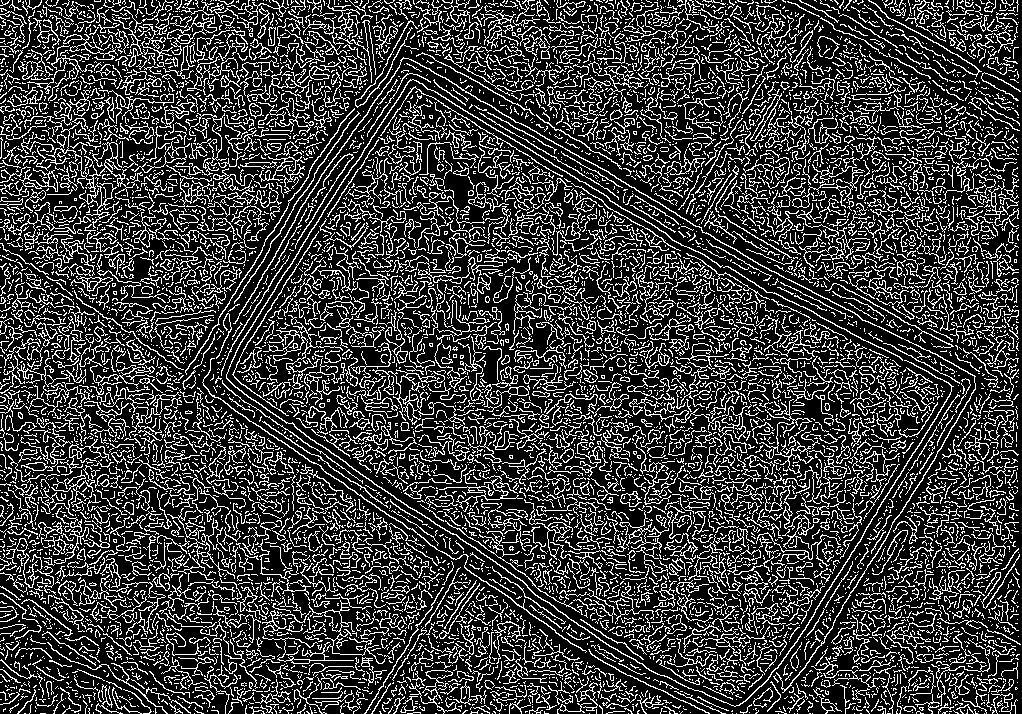}
\caption{}
\label{fig:subim1}
\end{subfigure}
\begin{subfigure}{0.23\textwidth \quad}
\includegraphics[width=1.65in,height=1.25in,keepaspectratio]{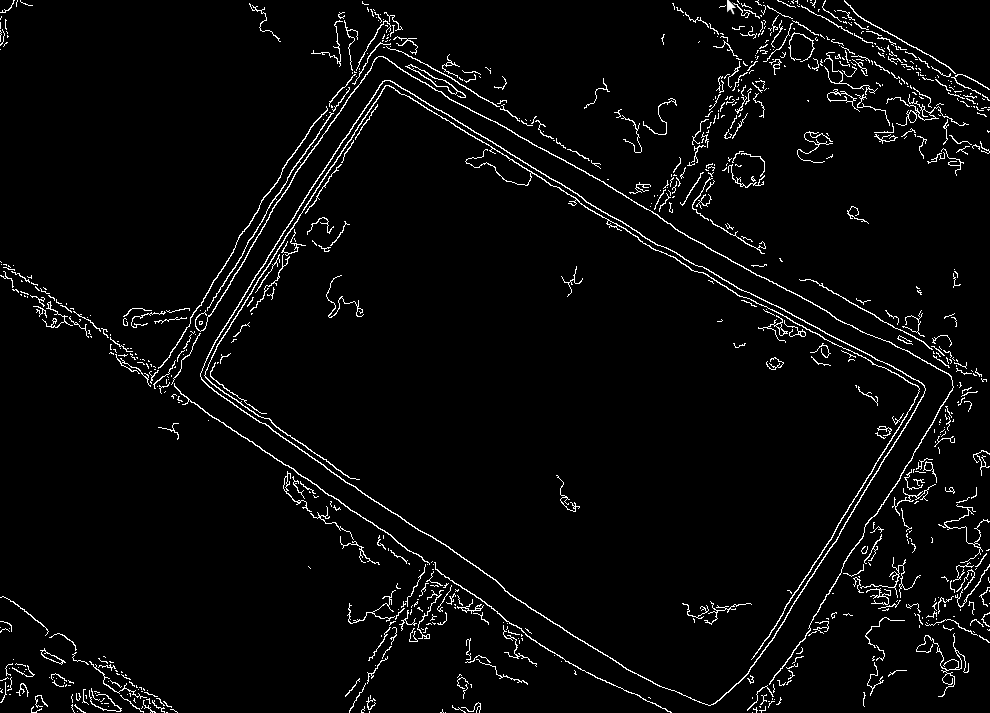}
\caption{}
\label{fig:subim2}
\end{subfigure}
\caption{(a)Edge detection before noise removal\\
         \- \hspace{0.85cm} (b)Edge detection after noise removal}
\label{fig:image1}
\end{figure}
Intensity values corresponding to desired edges are carried by quite less number of pixels which must be increased for getting a better quality detection. Histogram equalization technique is used to redistribute the intensities corresponding to detected edges which are our area of interest.The detected image after the noise removal steps is shown in the figure.1(b)
After Detecting the desired edges in the image , everything else other than window should be filtered out.For this purpose, Hough Probabilistic transform[8] is used.Each and every line in the image is represented in its r and $\theta$  form and by plotting the graph between these two quantities every point on that graph will represent a straight line(having corresponding r and $\theta$ value).By using a particular threshold votes (minimum number of intersections on the r-$\theta$ curve) every line is voted for its validity and by this method we can remove the lines other than window edges.Other parameters like minimum gap between two lines which should be maintained for the independent existence of those two lines and maximum width for a line to pass the test are also obtained using threshold values.These threshold values are decided by using trackbars on the image.The output of this method can be seen in figure.2(b). After getting the output from the hough transform, holes (less quality detection) in any edge might lead to a unclosed contour.For this purpose, dilation operation is applied which actually increases the width of the edge so that the potential holes will be filled up.
\begin{figure}[h]
\begin{subfigure}{0.2\textwidth}
\includegraphics[width=1.5in,height=1.25in,keepaspectratio]{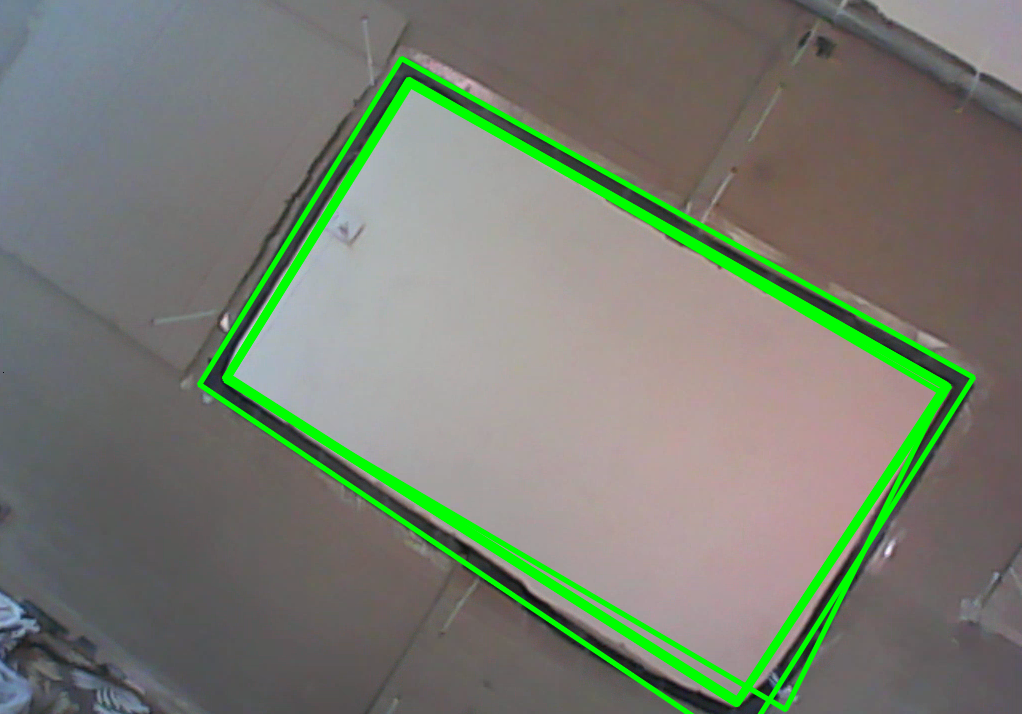}
\caption{}
\label{fig:subim1}
\end{subfigure}
\begin{subfigure}{0.24\textwidth \quad}
\includegraphics[width=1.5in,height=1.25in,keepaspectratio]{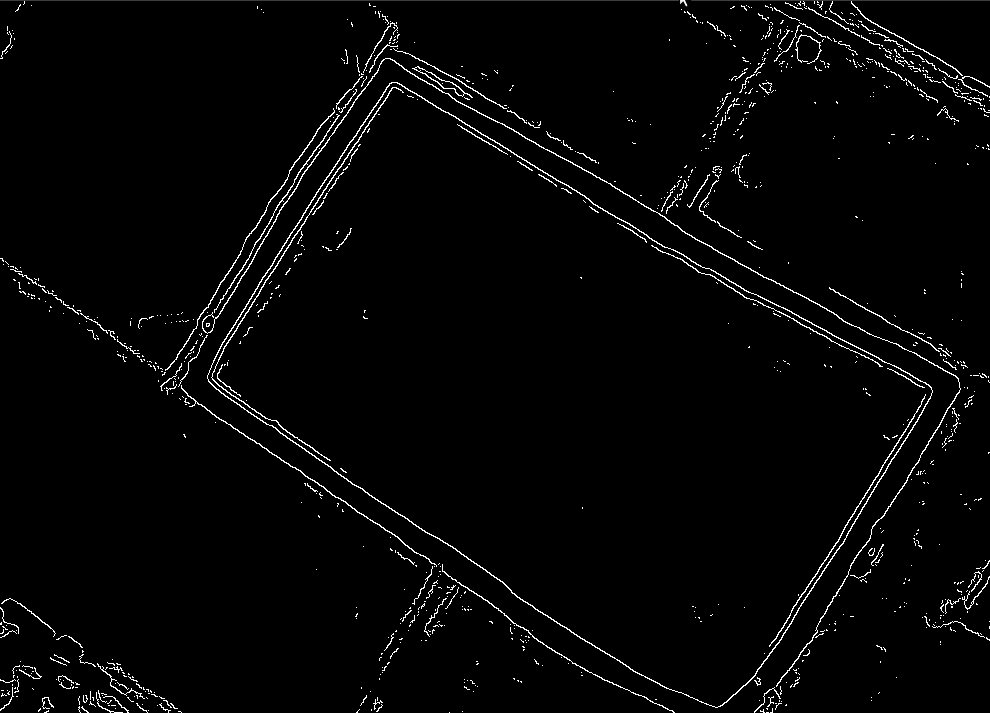}
\caption{}
\label{fig:subim2}
\end{subfigure}
\begin{subfigure}{0.2\textwidth }
\includegraphics[width=1.5in,height=1.25in,keepaspectratio]{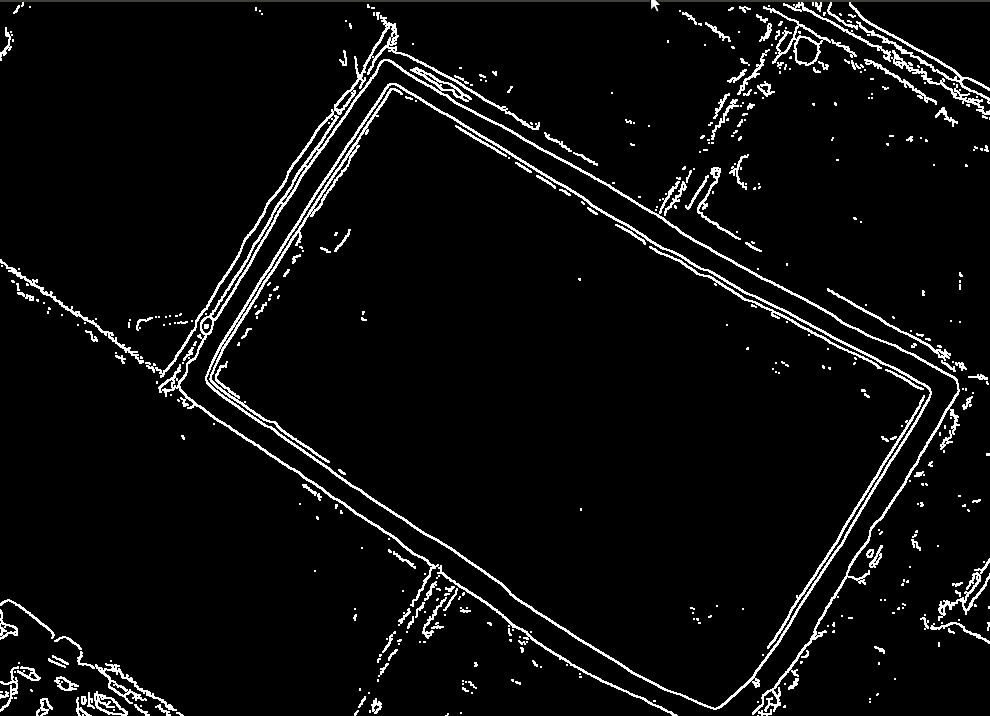}
\caption{}
\label{fig:subim3}
\end{subfigure}
\begin{subfigure}{0.23\textwidth \quad}
\includegraphics[width=1.5in,height=1.25in,keepaspectratio]{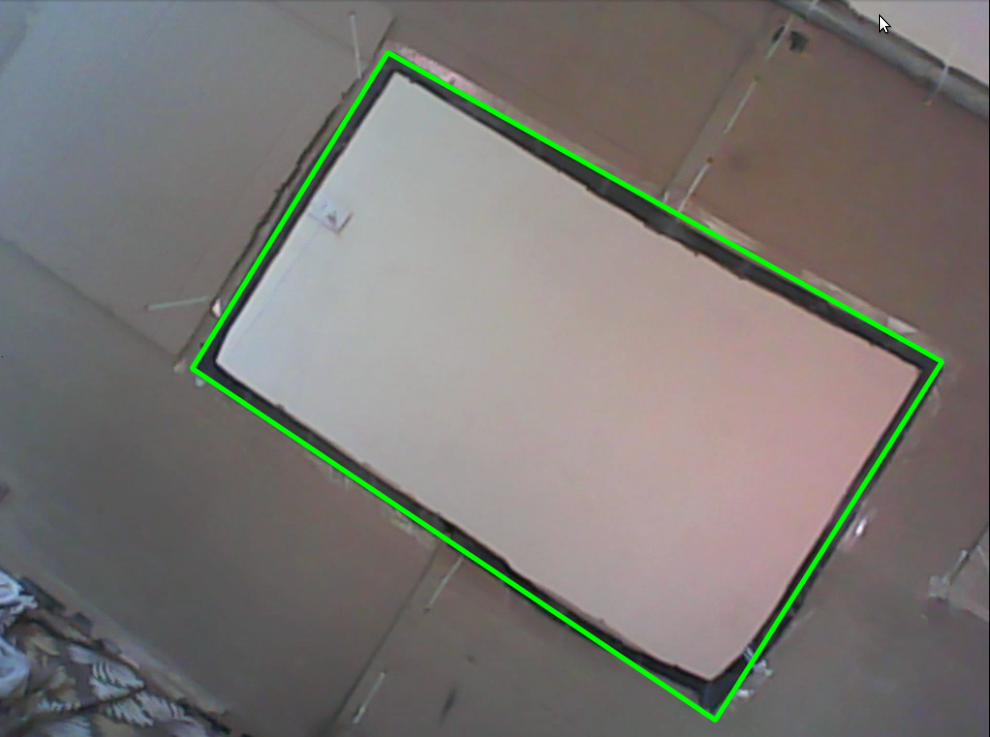}
\caption{}
\label{fig:subim4}
\end{subfigure}

\caption{(a)Bad detection because of noise and false positives\\
         \- \hspace{0.85cm} (b)Hough transform output\\
         \- \hspace{0.85cm} (c)Dilated output\\
         \- \hspace{0.85cm} (d)Correct robust window detection after all steps}
\label{fig:image2}
\end{figure}

\subsubsection{\textbf{Contour Selection}}
After getting the appropriate edges of the window, closed contours present in the image have to be detected and decided  whether it can be the target  window or not by applying some constraints.
 The vector of contours present in the image are selected and these are passed through an approximation step which generates a best fit approximated polynomial corresponding to that particular contour.
After getting approximate polynomial corresponding to a contour, the potential polynomials having the most probable chances of being window should be extracted by applying different constraints and conditions.These constraints and conditions are described below.

\textbf{Corners:}  There should be four corners in the detected contours for it to be a window.

\textbf{Area:} Area of the detected contour should be put in a certain range in order to remove the noisy contours.The min and max values of the threshold are decided according to the distance range within which UAV has to achieve its initial waypoint.

\textbf{Aspect-ratio:} The ratio of the width of the contour to the its height is the aspect ratio which should also be in a reasonable range so that the most of the distorted quadrilateral should be eliminated.

\textbf{Hull area:}  This is the area of the convex form of the contour detected.As the contour should be convex so this area should be almost equal to the contour area.

\textbf{Angle:} The angle between the edges should be almost 90 degrees as UAV will be in a particular yaw range.The threshold value of the angle will be decided according the maximum absolute yaw UAV will achieve in the task.

After getting the most potential contours, green rectangle is drawn on the image corresponding to the detected contour and centroid of the rectangle has been calculated and also drawn as yellow dot at the calculated pixel coordinates in the image.The Final window Detection is shown in the figure.2(d).\\
 \subsubsection{\textbf{Removal of False Positives}}
 While detecting the target window there are always false positives present in the surroundings which also qualify all the conditions for being a window ,but these are not our target window so all these false positives have to removed.

Histogram Comparison[9] technique is applied to remove these false positives.
In this method,histogram of the reference image( target Window image)and  the current image feed are compared to each other and Bhattacharyya distance is calculated which is low if the two images are almost same (if they have the same color composition) and high if they are not.By setting a threshold value on this distance, all the contours can be extracted which have this distance measure less than the threshold and all the other contours will be removed.
After removing the false detections and detecting the correct target window, the robust window detection is achieved.
\subsection{ \textbf{Camera Pose Estimation}}
Once the window is detected robustly , camera pose is needed in order to get the relative angle between UAV front plane and the window plane. Homographic Estimation[11] method is applied in order to get the camera poses.
\subsubsection{\textbf{Homography Matrix Calculation}}
The Homography matrix is the multiplication of the camera intrinsic matrix (K), the relative rotation matrix(R) and translation matrix(t) between the two views.Planar homography is defined as a projective mapping from one plane (object plane) to another (image plane). 
\begin{equation}
 \text{$H$ =$\left[\begin{array}{ccc}
fs_x & 0 & c_x  \\
0 & fs_y & c_y   \\
0 & 0 & 1\end{array}\right]\left[\begin{array}{cccc}
r_{11}& r_{12}  & r_{13} & T_x\\
 r_{21}& r_{22}  & r_{23} & T_y \\
r_{31}& r_{32}  &  r_{33} & T_z\end{array}\right]$
}
\end{equation}
Using homogeneous coordinates, the relationship between two planes can be expressed as 
\begin{equation} 
\text{  $\left[ \begin{array}{ccc}
x_i  \\
y_i  \\
1  \end{array} \right]= sH\left[\begin{array}{ccc}
X_w   \\
Y_w   \\
Z_w   \\
1
\end{array}\right]$} 
\end{equation}\\
$s$ is an arbitrary scale factor because homograph is only defined up to a scale factor.$x_i,y_i$are in image pixels and $X_w , Y_w , Z_w $ are the world co-ordinates.
The projective transformation from one plane to another plane can be written as -

\begin{equation}
 \text{$ \left[ \begin{array}{ccc}
x_1  \\
x_2  \\
x_3  \end{array} \right]$= $M\left[\begin{array}{ccc}
X_w   \\
Y_w   \\
Z_w   \\
1
\end{array}\right]$=$M_{int}M_{ext}\left[\begin{array}{ccc}
X_w   \\
Y_w   \\
0   \\
1
\end{array}\right]$}
\end{equation}
\\
where M is the projection matrix and
since the object is planar, the world coordinate $Z_w$ of all 
corner points can be set to zero.\\
\\
\begin{equation}
\text{$ \left[ \begin{array}{ccc}
x_i  \\
y_i  \\
1\end{array} \right]$= $sH\left[\begin{array}{ccc}
X_w   \\
Y_w   \\
Z_w   \\
1
\end{array}\right]$=$sM_{int}M_{ext}\left[\begin{array}{ccc}
X_w   \\
Y_w   \\
0   \\
1
\end{array}\right]$
\\} 
\end{equation}
=$s \left[\begin{array}{ccc}
fs_x & 0 & c_x  \\
0 & fs_y & c_y   \\
0 & 0 & 1\end{array}\right]\left[\begin{array}{cccc}
r_{11}& r_{12}   & T_x\\
 r_{21}& r_{22}  & T_y \\
r_{31}& r_{32}   & T_z\end{array}\right]\left[\begin{array}{cc}
 X_w  \\
 Y_w  \\
 1
\end{array}\right]$
\\
The homography matrix H that maps a planar object’s points onto the image is described completely by a 3x3 matrix.
\begin{equation}
\text{ $\left[ \begin{array}{ccc}
x_i  \\
y_i  \\
1  \end{array} \right]= sH\left[\begin{array}{ccc}
X_w   \\
Y_w   \\ 
1
\end{array}\right]$}
\end{equation}\\
The above equation is used to compute the homography matrix for each view.Taking intrinsic parameters constant(depends only on the camera), the homography matrix is calculated continuously for each video frame with respect to the reference target frame.
\subsubsection{\textbf{Extracting Rotation Matrix}}
From equation $(1)$, in order to obtain the rotation matrix from the homography matrix,camera intrinsic parameter will be needed.The Intrinsic parameters of a camera can be calculated using camera calibration.\\
\quad \textbf{Intrinsic parameters :}$f{s_x},f {s_y},c_x,c_y$\\
3-D geometry $c_x,c_y$ is the principal point in pixels ,$s_x,s_y$ are the pixel size in pixels per mm or inches,f is the focal length in mm or inches, $\sigma$ is the aspect ratio $\frac{s_x}{s_y}$(usually 1.0 for a square pixel camera).These parameters are used collectively as Intrinsic camera matrix $K$.\\
\begin{equation}
\text{$K$ =
$\left[\begin{array}{ccc}
 fs_x& 0  & c_x\\
 0& fs_y  & c_y \\
0& 0 &  1\end{array}\right]$}
\end{equation}

\textbf{Extrinsic Parameter:}Values of Rotational $(R)$ and Translational $(T)$ Matrices are considered as Extrinsic parameters of a camera.These are also the pose of that camera.
Rotation matrix can be represented as rotation about the three individual axes.
\begin{equation}
\text{$R$ =\small{
$\left[\begin{array}{ccc}
 r_{11}& r_{12}  & r_{13}\\
 r_{21}& r_{22}  & r_{23} \\
r_{31}& r_{32}  &  r_{33}\end{array}\right]$}}
\end{equation}

\resizebox{\linewidth}{!}{%
$=\left[\begin{array}{cccc}
 1& 0  & 0 & 0\\
 0 & \cos{\theta}  & -\sin{\theta} & 0 \\
0 & \sin{\theta}  & \cos{\theta} & 0\\
0 & 0  & 0 & 1
\end{array}\right]\left[\begin{array}{cccc}
  \cos{\phi}  & 0 & \sin{\phi} & 0 \\
 0 & 1  & 0 & 0 \\
 -\sin{\phi}  & 0 &\cos{\phi} & 0\\
0 & 0  & 0 & 1
\end{array}\right]\left[\begin{array}{cccc}
  \cos{\psi}  & \sin{\psi} & 0 & 0 \\
 \sin{\psi} &\cos{\psi}   & 0 & 0 \\
 0 & 0  & 1 & 0 \\
0 & 0  & 0 & 1
\end{array}\right]$%
}
\begin{equation}
T=
\left[\begin{array}{ccc}
T_x  \\
T_y  \\
T_z   
\end{array}\right]
\end{equation}

The three angles ,which are the Euler angles(the relative angle between the camera plane and the reference window plane) are calculated using the following formulas.
\\ 

$\theta$ = $atan2(r_{32},r_{33})$\\
$ \phi $  =  $atan2(-r_{31},\sqrt{r_{32}^2+{r_{33}}^2})$\\
$\psi$   = $atan2(r_{21},r_{11})$\\
 These angles are used in navigation scheme of UAV.
By using these angles , the deviation of UAV from its desired position will be known and according to these angles left/right and yaw commands will be given as shown below in the figure.$3$ The obtained angles can also be seen in a video[10].
\begin{figure}[h]
\begin{subfigure}{0.2\textwidth}
\includegraphics[width=1.7in,height=1.25in,keepaspectratio]{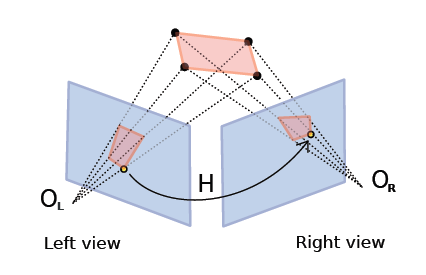}
\caption{}
\label{fig:subim1}
\end{subfigure}
\begin{subfigure}{0.23\textwidth \quad}
\includegraphics[width=1.65in,height=1.25in,keepaspectratio]{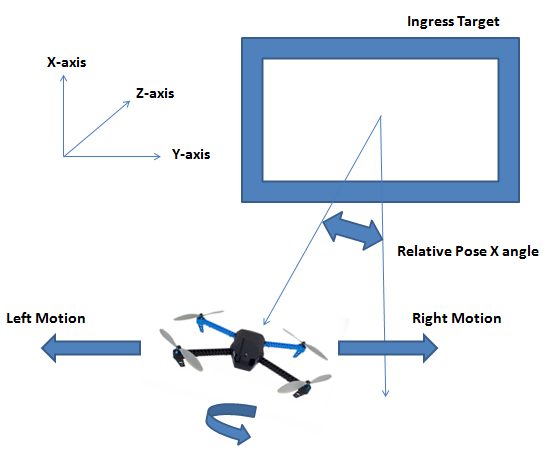}
\caption{}
\label{fig:subim2}
\end{subfigure}
\caption{(a) Homography Between Two planes\\
         \- \hspace{0.85cm} (b)Navigation of UAV using relative pose angles}
\label{fig:image1}
\end{figure}
\subsection{Navigation Scheme}
The scheme used to navigate the UAV and to ingress through the target window is described below -
\begin{itemize}
\item Initially UAV will do only yaw motion while taking off,in order to search the target window.Once the target window is detected correctly,it will maintain that height and will stop yaw motions.  
\item It calculate the relative angle between its current position and the desired position( where its forward direction will be perpendicular to the window plane and it will be aligned to the center of the window).
 \item It will start moving in that particular direction in which the relative angle(obtained in the previous step) will reduce.It will only perform motions in Y plane(left/right motions) and also yaw motions in order to keep detecting the whole window properly.
\item It will adjust its height continuously according the height of window center .
\item Once it reach the desired position ,it will start motion towards the window (motion in X plane)and will perform previous steps iteratively.
\item Whenever it will not detect window , by default , it start moving away from the window in order to get it in the frame and will start detection again.
\end{itemize}
\section{Testing and results}
This navigation scheme is tested in a Simulation environment Gazebo[12], a well designed simulation environment which allows efficient and accurate simulations even in complex indoor and outdoor environments.Gazebo uses ROS[13] methods in order to connect its different components. A world(simulation environment) was created using a UAV model, a window and a house model.
As pointed in the navigation scheme part, this visual based navigation is done iteratively.The motions in Y-axis(left/right motions) and yaw motions will be performed in steps.These step sizes are sufficiently small in order to keep checking the validity of relative angles. This makes the navigation a little slow which can be resolved in future by applying the algorithm more efficiently.
The system is able to generate highly accurate poses which are confirmed and illustrated by the convergence of the relative angle as shown in fig-4(a).To confirm the quality of the reconstruction results, opening width estimation results and comparison results of left side and right side width of  ingress target are added in the ingress experiment as shown in the fig-4(b).

In this algorithm the commands will only be given for a valid angle so that the fraction of the total deviation covered till then after each command ,would not go in vain.   
This algorithm was tested under conditions specified below-
\begin{itemize}
\item The horizontal axes of the camera mounted on the quadrotor is perpendicular to the vertical axes of quadrotor.
\item The intensity values of the other side of the window are different from that of the neighboring regions.
\end{itemize}
The first assumption affects the perceived shape of the ingress target.Second assumption affects the detection part which actually affects the accuracy of Ingress.
But in our case quadrotor maintain its height at the height of window center using altitude hold, so the first assumption would be approximately correct i.e the the vertical axis of the quadrotor would actually be perpendicular to the ground. 

To confirm the correctness of the navigation scheme,extra measurement is also added which is opening width of the detected window. It is plotted with respect to motion in Y - direction .This plot checks whether quadrotor will be at its maximum value at the time of ingress or not in a typical test run. Apart from these quantities the relative angle with respect to the motion in Y-axis is also plotted.
\begin{figure}[h]
\begin{subfigure}{0.2\textwidth}
\includegraphics[width=1.7in,height=1.25in,keepaspectratio]{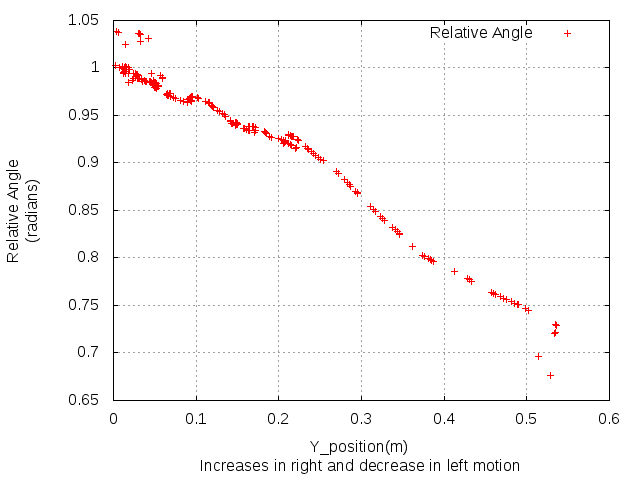}
\caption{}
\label{fig:subim1}
\end{subfigure}
\begin{subfigure}{0.23\textwidth \quad}
\includegraphics[width=1.65in,height=1.25in,keepaspectratio]{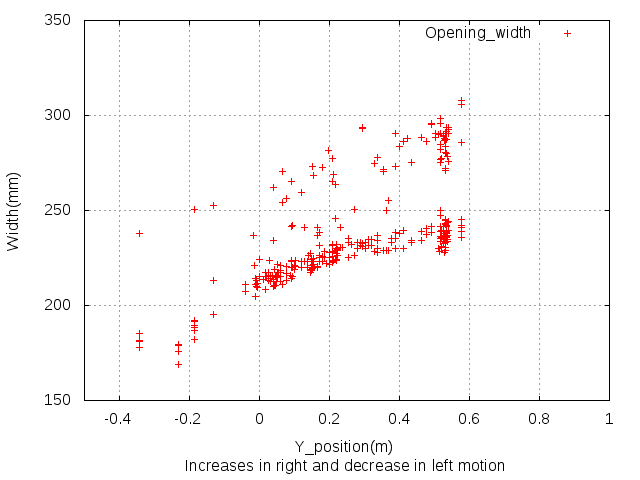}
\caption{}
\label{fig:subim2}
\end{subfigure}
\caption{(a)Convergence test of the estimated relative angle\\
         \- \hspace{0.85cm} (b)Opening width estimation for the ingress target}
         \- \hspace{0.65cm}(Y-position increases in  right and decreases in left motion)
\label{fig:image1}
\end{figure}
\section{Conclusion and Future work}
This paper proposed a method to robustly detect the window in the surrounding using basic image processing techniques and efficient distance measure.Furthermore, a navigation scheme which incorporates the detection method for performing navigation task was also proposed.Homographic estimation was used to correctly obtain relative angles/poses.These relative angles were used in the navigation scheme.This navigation scheme is one of the vision-based, experimentally tested algorithms for small UAVs, which uses a single monocular camera and requiring no artificial labeling or external computation and thus enabling ingress into a building.\\
A challenge for this type of vehicle when maneuvering close to the structure or in wind gusts is the incorporation of a dynamic controller to consider changes in vehicle dynamics caused by aerial disturbances. In particular, exploration of  this aspect of the problem for  performing autonomous ingress maneuvers through small openings is remaining as a future work.The visual algorithm used under certain assumptions like- The intensity values of the other side of the window are different from that of the neighboring regions.Making visual algorithms independent of these assumptions is also remaining as a future work. 

\ifCLASSOPTIONcaptionsoff
  \newpage
\fi

\end{document}